\documentclass[10pt,twocolumn,letterpaper]{article}

\usepackage{iccv}
\usepackage{times}
\usepackage{epsfig}
\usepackage{graphicx}
\usepackage{amsmath}
\usepackage{amssymb}
\usepackage{pifont}
\usepackage{bm}
\usepackage{balance}
\usepackage{booktabs}
\usepackage{multirow}
\usepackage{threeparttable}
\usepackage{bbding}
\usepackage[accsupp]{axessibility}  % Improves PDF readability for those with disabilities.

\usepackage[pagebackref=true,breaklinks=true,letterpaper=true,colorlinks,bookmarks=false]{hyperref}

\iccvfinalcopy % *** Uncomment this line for the final submission

\def\iccvPaperID{2909} % *** Enter the ICCV Paper ID here

\ificcvfinal\fi

\begin{document}

%%%%%%%%% TITLE
\title{ A Baseline Framework for Part-level Action Parsing and Action Recognition \\
\normalsize 2nd place solution for Kinetics-TPS Track in ICCV DeeperAction Workshop 2021
}

\author{Xiaodong Chen$^1$\thanks{This work was done when Xiaodong Chen was an intern at JD AI Research. }\quad
Xinchen Liu$^2$ \quad
Kun Liu$^2$ \quad
Wu Liu$^2$ \quad
Tao Mei$^2$ \\
$^1$University of Science and Technology of China, Hefei, China \\ 
$^2$JD AI Research, Beijing, China \\
{\tt\footnotesize cxd1230@mail.ustc.edu.cn,liuxinchen1@jd.com,liu\_kun@bupt.edu.cn,liuwu@live.cn,tmei@live.com}\\ %\vspace{-1mm}
% {\tt\small\texttt{xzhang@ee.ryerson.ca, zyd73@ustc.edu.cn, tmei@live.com}}
}

% \author{Xiaodong Chen\\
% University of Science and Technology of China\\
% Hefei, China\\
% {\tt\small cxd1230@mail.ustc.edu.cn}
% \and
% Xinchen Liu\\
% AI Research of JD.com\\
% Beijing, China\\
% {\tt\small liuxinchen1@jd.com}
% \and
% Wu Liu\\
% AI Research of JD.com\\
% Beijing, China\\
% {\tt\small liuwu1@jd.com}
% \and
% Xiao-Ping Zhang\\
% Ryerson University\\
% Toronto, Canada\\
% {\tt\small liuxinchen1@jd.com}
% \and
% Yongdong Zhang\\
% University of Science and Technology of China\\
% Hefei, China\\
% {\tt\small zyd73@ustc.edu.cn}
% Tao Mei\\
% AI Research of JD.com\\
% Beijing, China\\
% {\tt\small tmei@jd.com}
% }

\maketitle
\ificcvfinal\thispagestyle{empty}\fi

%%%%%%%%% ABSTRACT
\begin{abstract}
This technical report introduces our 2nd place solution to
Kinetics-TPS Track on Part-level Action Parsing in ICCV DeeperAction Workshop 2021. 
Our entry is mainly based on YOLOF~\cite{conf/cvpr/ChenWYZC021} for instance and part detection, HRNet~\cite{conf/cvpr/0009XLW19} for human pose estimation, and CSN~\cite{conf/iccv/TranWFT19} for video-level action recognition and frame-level part state parsing. 
We describe technical details for the Kinetics-TPS dataset, together with some experimental results. 
In the competition, we achieved 61.37\% mAP on the test set of Kinetics-TPS.
\end{abstract}

%%%%%%%%% BODY TEXT
\section{Method}
\label{sec:method}
\subsection{Overall Framework}
\label{subsec:framework}
In the Kinetics-TPS competition, one of the most challenging tasks is body part detection.
Our approach innovatively takes the results of pose estimation as position coding, which significantly improves the accuracy of the part-level detection network. 
Meanwhile, by observing the Kinetics-TPS dataset, we transform the frame-level part state parsing into video-level action recognition~\cite{conf/aaai/LiuLGTM18}.

\begin{figure*}
\begin{center}
\includegraphics[width=0.9999\linewidth]{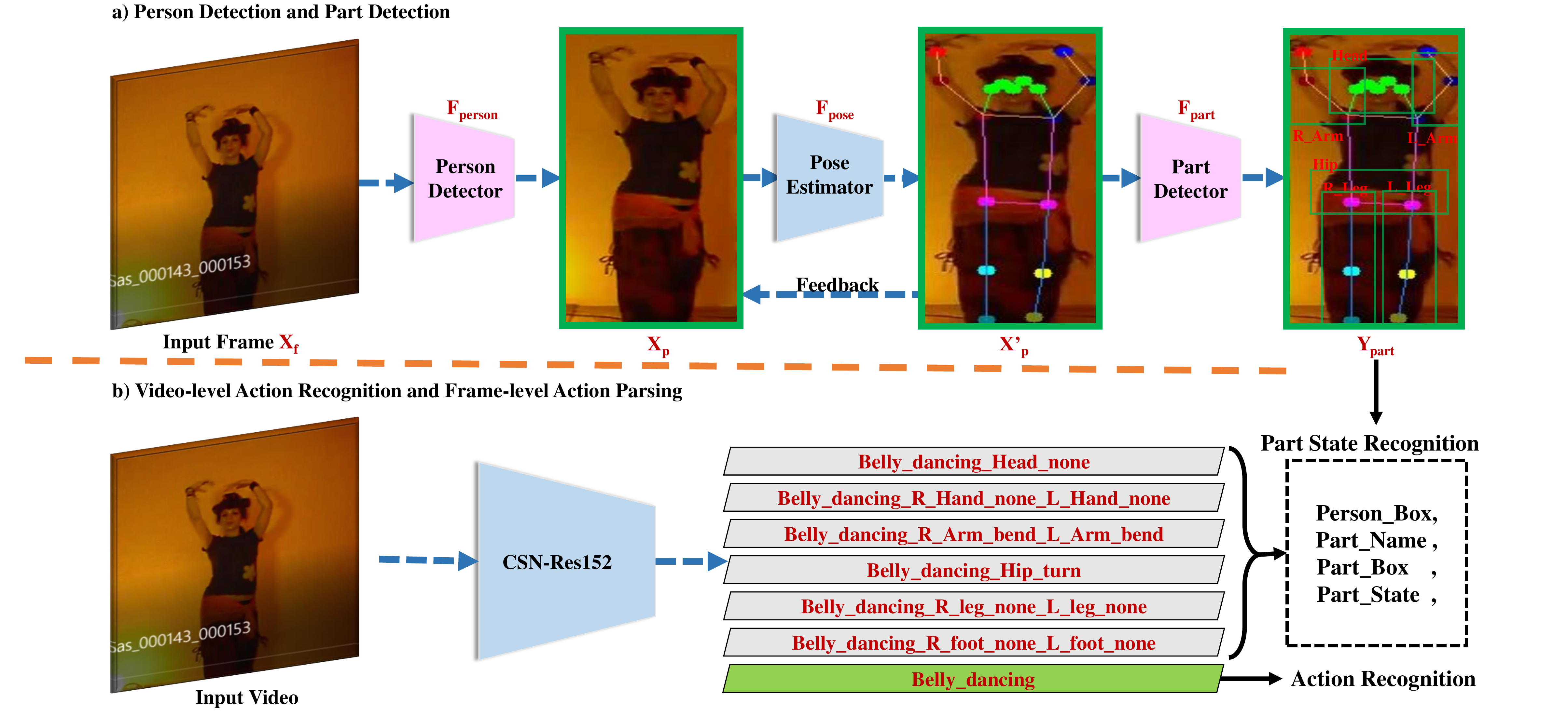}
\end{center}
   \caption{The overall architecture of our method.}
\label{fig:figure1}
\end{figure*}

The overall framework for the competition is shown in Figure~\ref{fig:figure1}. 
It includes three modules for instance and part detection, video-level action recognition, and frame-level part state parsing, 
Next, we will introduce each module in our framework.

%Figure 1

\subsection{Pose-guided Part Detection}
\label{subsec:detection}
To the best of our knowledge, body part detection is an unprecedented task in traditional detection tasks. 
Different from traditional object detection tasks, body part detection has a strong prior to the human body structure.
For example, normally people only have one left foot and one right foot. 
This strong prior is very common in human pose estimation and has been well and widely utilized. 
In order to maximize the usage of this strong prior, we use the output of a human pose estimator to help the detector better predict the boxes of human parts.
Therefore, we propose a pose-guided part detection method, which is shown in the top half of Figure~\ref{fig:figure1}. 

In detail, the person detector $F_{person}$ first extracts the bounding box of a person $X_p$ from the input frame $X_f$.
Then the pose estimator $F_{pose}$ takes $X_p$ as the input and outputs the position of keypoints $K_{p}$ of the person.
After that, the keypoints $K_{p}$ are drawn by dots of different colors $G$ on the original person image of $X_p$ to generate an augmented person image $X'_p$.
By this means, we can increase the appearance difference between different body parts and facilitate the learning of body parts detector $F_{body}$. 
Finally, the part detector $F_{body}$ is implemented to localize the part boxes $Y_{part}$ of each body part.
This process can be formulated by
\begin{equation}
X_{p} = F_{person}(X_{f}),
\label{equ:equ1}
\end{equation}
\vspace{-4mm}
\begin{equation}
K_{p} = F_{pose}(X_{p}),
\label{equ:equ2}
\end{equation}
\vspace{-4mm}
\begin{equation}
X'_{p} = X_{p} + G(K_{p}),
\label{equ:equ3}
\end{equation}
\vspace{-4mm}
\begin{equation}
Y_{Part} = F_{body}(X'_{p}).
\label{equ:equ4}
\end{equation}

In addition, we also fine-tune the person detection box with the results of the pose estimator.
In a nutshell, the pose estimator has the ability to predict the possible human keypoints outside the person box, and we fine-tune the detected person box until all possible human keypoints are included.

\subsection{Part State Parsing and Action Recognition}
\label{subsec:parsingtorecongnition}
Part state parsing is similar to the spatial-temporal action detection task. 
However, we find this problem can be transformed into a simpler video understanding task due to the overwhelming ``Long Tail Effect'' in the Kinetics-TPS dataset.
For example, in the video of Capoeira, we just need to predict ``None'' for the heads in every frame and easily achieve 96.5\% frame-level part state accuracy. 
To take advantage of the significant ``Long Tail Effect'', as shown in the bottom half of Figure~\ref{fig:figure1}, we label each video with 6 part-level labels on the basis of the original video-level label.
The part-level label consists of three parts: video-level action, body part, and the most frequently frame-level action of the body part. 
As the example shown in Figure~\ref{fig:figure1}, ``belly\_dancing\_Head\_none'' means the video-level action of this video is ``belly\_dancing'', and the most frequently frame-level action of the ``head'' in this video is ``none''. 
Through this transformation, we can directly apply seven (six models for frame-level action prediction and one for video-level prediction) individual Action Recognition Networks, such as CSN~\cite{conf/iccv/TranWFT19}, to predict each label without any other models related to the spatial-temporal action detection tasks.

\section{Experiments}
\label{sec:experiments}

\subsection{Implementation Details}
\label{subsec:details}

%Table 1
\begin{table*}[t]
\begin{center}
    \footnotesize
    \begin{threeparttable}
    \begin{tabular}{l|c|c|c}
    \toprule
    Submission & Video Acc (\%) & Test Score (\%) & Details\\
    \midrule
    1 & 86.57 & 45.11 & Baseline.\\
    2 & 96.78 (\textbf{+10.21}) & 50.22 (\textbf{+5.11})  & Freezing BN layers in the CSN model and using the results of CSN-ensemble.\\
    3 & 96.78 & 58.31 (\textbf{+8.09}) & Using the pose-guided part detector in subsection~\ref{subsec:detection}.\\
    4 & 96.78 & 58.66 (\textbf{+0.35}) & Using the pose estimator to fine-tune the person detection box.\\
    5 & 96.78 & 61.10 (\textbf{+2.44}) & Converting the frame-level part state parsing problem into video-level action recognition.\\
    6 & 96.78 & 61.37 (\textbf{+0.27}) & Using CSN-ensemble model for Part State Parsing. \\
    \bottomrule
    \end{tabular}
    \end{threeparttable}
    \caption{Ablation results of different submissions on the Kinetics-TPS testing set. ``Video Acc'' in the second column refers to the top-1 video-level action recognition accuracy. ``Test Score'' in the third column refers to the final scores of the submissions.}
    \label{tab:table1}
\end{center}
\end{table*}

%Table 2
\begin{table*}[t]
\begin{center}
    \footnotesize
    \begin{threeparttable}
    \begin{tabular}{l|c|cccccc}
    \toprule
    Models & Pre-train Datasets  & Head (\%) & Hand (\%) & Arm (\%) & Hip (\%) & Leg (\%) & Foot (\%)\\
    \midrule
    ip-CSN-152 & IG-65M &  92.16 (1.413) & 65.41 (1.314) & 60.08 (1.0) & 80.99 (1.906)  & 65.76 (3.515)  & 65.84 (1.576)  \\
    ir-CSN-152 & Sports-1M  & 86.31 (1.0) & 59.48 (1.0) & 55.60 (0.0) & 77.52 (1.0)  & 59.30 (1.0)  & 61.24 (1.0)  \\
    \midrule
    CSN-ensemble & - &  92.41 & 65.94 & 60.08  & 82.36  & 66.66 & 66.70  \\
    \bottomrule
    \end{tabular}
    \end{threeparttable}
    \caption{Top-1 accuracy on Kinetics-TPS. The last six columns mean the video-level accuracy of different body parts calculated on the validation set. We set the number of clips as 7 during the inference stage. The numbers in parentheses following the accuracy represent the fusion weights at the final CSN-ensemble.}
    \label{tab:table2}
\end{center}
\end{table*}

\textbf{Dataset.} 
For this year's challenge of part-level action parsing, the dataset named Kinetics-TPS provides 7.9M annotations of 10 body parts, 7.9M part state (i.e., how a body part moves) of 74 body actions, and 0.5M interactive objects of 75 objects in the video frames of 24 human action classes. 
Kinetics-TPS contains 3,809 training videos (4.96GB in size) and 932 test videos (1.26GB in size). 
For the needs of this challenge, we divide the training videos into the training set containing 3000 videos and the valuation set containing 809 videos. 
Following the guidelines of the challenge, we set HUMAN\_IOU\_THRESH as 0.5 and set PART\_IOU\_THRESH as 0.3 in frame-level action prediction. 
Please refer to \href{https://github.com/xiadingZ/Kinetics-TPS-evaluation}{Kinetics-TPS-evaluation} for detailed calculation of evaluation indicators.

\textbf{Detector and Pose Estimator.} 
For person detector and part detector on keyframes, we adopted the YOLOF~\cite{conf/cvpr/ChenWYZC021}, which is an anchor-free model with a ResNet-101~\cite{conf/cvpr/HeZRS16} backbone. 
The model is pre-trained on the COCO dataset~\cite{conf/eccv/LinMBHPRDZ14} and then fine-tuned on Kinetics-TPS. 
The final models obtain 93.8 AP@50 in the person category and 79.7 AP@50 in the 10 body parts categories on the Kinetics-TPS validation set.
For the pose estimator, we directly adopted the HRNet-w48~\cite{conf/cvpr/0009XLW19} pre-trained on COCO~\cite{conf/eccv/LinMBHPRDZ14} to extract the keypoints of each person without any fine-tuning.

\textbf{Action Recognition Network.} 
We use the CSN networks~\cite{conf/iccv/TranWFT19} as the backbone in our action recognition and action parsing framework.
We use ip-CSN-152 and ir-CSN-152 implementations that pre-trained on IG-65M~\cite{conf/cvpr/GhadiyaramTM19} and Sports-1M~\cite{conf/cvpr/KarpathyTSLSF14} datasets respectively with input sampling $T \times \tau = 32 \times 2$.
In particular, we freeze the Batch Normalization (BN) layer in the backbone during fine-tuning on Kinetics-TPS. 

\textbf{Training.} 
The detection model and action recognition models are trained separately.
Each model is trained in an end-to-end fashion.
In detail, we train the YOLOF detector using SGD with a mini-batch size of six on four V100 GPUs and train it for 24 epochs with a base learning rate of 0.01, which is decreased by a factor of 10 at epochs 16 and 22. 
We perform linear warm-up~\cite{journals/corr/GoyalDGNWKTJH17} during the first 1800 iterations.
For the CSN model, we train it using SGD with a mini-batch size of four on four V100 GPUs for 58 epochs with a base learning rate of 0.00008, which is decreased by a factor of 10 at epochs 32 and 48. 
We perform linear warm-up~\cite{journals/corr/GoyalDGNWKTJH17} during the first 16 iterations. 
By default, we use weight decay of 0.0001 and Nesterov momentum of 0.9 for all models.

\textbf{Inference.} 
During testing, we extract the top-10 results from the person detector and the top-1 results of each body part from the part detector. 
For the video understanding task and action parsing task, we set the number of clips as seven for each video at test time and scale the shorter side of input frames to 256 pixels.

%Table 3
\begin{table}[t]
\begin{center}
    \footnotesize
    \begin{threeparttable}
    \begin{tabular}{l|c|c|c}
    \toprule
    Models & Pre-train Datasets & Freeze BN & Video Acc (\%)\\
    \midrule
    ip-CSN-152 & IG-65M & \checkmark & 96.46 (7.0)  \\
    ir-CSN-152 & Sports-1M & \checkmark & 92.60 (1.0)  \\
    \midrule
    CSN-ensemble & - & - & 96.78 \\
    \bottomrule
    \end{tabular}
    \end{threeparttable}
    \caption{Top-1 video-level action recognition accuracy on Kinetics-TPS. The number in parentheses following the accuracy in the table represents the fusion weights at the final CSN-ensemble. In this table, \checkmark means freezing all BN layers in the backbone network.}
    \label{tab:table3}
\end{center}
\end{table}

%TODO
\subsection{Main Results}
\label{subsec:results}
We present our results on Kinetics-TPS in Table~\ref{tab:table1}.
The ``video acc'' in the second column refers to the top-1 video-level action recognition accuracy, while the ``test score'' in the third column refers to the final score of the commit.

The baseline backbone for video-level action recognition is the ip-CSN-152 model pre-trained on the IG-65M dataset without freezing BN layers.
As for the part state parsing task of baseline commit, we directly predict the frame-level action label of each body part to be the action that appears most frequently in each class in the trainset without any deep learning models. 
For instance, we predict the action of ``hip'' in every frame of ``belly\_dance'' videos as ``turn'' for the reason that the most frequent action of ``hip'' in videos about ``belly\_dance'' is ``turn''. The simplest baseline has only 45.11 mAP.

With the help of freezing BN layers in the CSN model and combining predictions of two models with two different backbones, as shown in Table~\ref{tab:table3}, our second commit achieves a significant enhancement of {+5.11} mAP in the final score.

Moreover, adding the aforementioned pose-guided part detector and fine-tuning the person detection box gives a total boost of {+8.44} mAP.

For the final two commits, we convert the frame-level part state parsing problem into video-level action recognition as described in subsection~\ref{subsec:parsingtorecongnition}. This measure further brings considerable improvement in performance ({+2.71} mAP). As shown in Table~\ref{tab:table2}, the video-level accuracy of different body parts is calculated on the validation set because the real frame-level action label of the test set is not obtainable even after the end of the challenge.

%Table 4
\begin{table}[t]
\begin{center}
    \footnotesize
    \begin{threeparttable}
    \begin{tabular}{l|c|c|cc}
    \toprule
    Model & Pre-train & Pose & $AP$ (\%) & $AP_{50}$ (\%) \\
    \midrule
    $YOLOF_{person}$ & COCO & \ding{55}    & 74.60 & 93.40 \\
    $YOLOF_{person}$ & COCO & \checkmark & 74.80 (\textbf{+0.20}) & 93.80 (\textbf{+0.40})  \\
    $YOLOF_{part}$   & COCO & \ding{55}    & 36.40 & 53.10 \\
    $YOLOF_{part}$   & COCO & \checkmark & 57.10 (\textbf{+20.7}) & 79.70 (\textbf{+26.6}) \\
    \bottomrule
    \end{tabular}
    \end{threeparttable} 
    \caption{Effect of pose estimator. In this table, \checkmark means using pose estimator for the detector.}
    \label{tab:table4}
\end{center}
\end{table}

%Table 5
\begin{table}[t]
\begin{center}
    \footnotesize
    \begin{threeparttable}
    \begin{tabular}{l|c|c|c}
    \toprule
    model & pre-train dataset & BN & Video Acc (\%) \\
    \midrule
    ir-CSN-152 & scratch & \ding{55}     & 57.50 \\
    ir-CSN-152 & IG-65M  & \ding{55}     & 85.60 (\textbf{+28.10}) \\
    ir-CSN-152 & IG-65M  & \checkmark    & 95.49 (\textbf{+9.89}) \\
    \bottomrule
    \end{tabular}
    \end{threeparttable}
    \caption{Pre-training on IG-65M dataset and Freezing BN Layers. In this table, \checkmark means freezing all BN layers in the backbone network}
    \label{tab:table5}
\end{center}
\end{table}

\subsection{Ablation Experiments}
\label{subsec:ablation}

\textbf{Effect of Adding Pose Estimator.} 
We investigate the effect of the pose estimator on detection mAP. 
For person detector and part detector, we train the same model YOLOF with human boxes and body parts boxes. 
As shown in Table~\ref{tab:table4}, adding the pose estimator brings consistent $AP$ and $AP_{50}$ increases for these two models. 
More specifically, equipped with the pose estimator, our $YOLOF_{part}$ model achieves a significant enhancement of $+26.6$ $AP_{50}$ on the Kinetics-TPS dataset.

\textbf{Pre-training and Freezing.} 
We quantitatively measure the importance of freezing BN layers and using pre-train datasets in Table~\ref{tab:table5}. 
All experiments are based on the ir-CSN-152 model. 
It is also worth noting that we set the number of clips as seven at test time.
From the results, we can see that the pre-training on the IG-65M dataset can bring 28.1\% improvement over training from scratch.
Moreover, freezing BN layers can further improve the performance by 9.89\%.

\section{Conclusion}
This paper presents a baseline framework for part-level action parsing and action recognition and won 2nd place in Kinetics-TPS Track of ICCV DeeperAction Workshop 2021. In our baseline framework, the Pose-guided Part Detection is one of the first attempts toward body part detection and brings considerable improvement in the final score (+8.09\%). Meanwhile, Converting the frame-level part state parsing problem into video-level action recognition gives a total boost of +2.71\% in the final score. With these two contributions, we provide a strong baseline for the spatial-temporal action parsing task.

%\newpage
{\small
\balance
\bibliographystyle{ieee_fullname}
\bibliography{iccv2021}
}

\end{document}